\documentclass{article}

\usepackage[preprint]{airov24}

\usepackage[utf8]{inputenc} 
\usepackage[T1]{fontenc}    
\usepackage{hyperref}       
\usepackage{url}            
\usepackage{booktabs}       
\usepackage{amsfonts}       
\usepackage{nicefrac}       
\usepackage{microtype}      
\usepackage{xcolor}         

\usepackage{biblatex}
\usepackage{graphicx}

\title{Enhancing Historical Image Retrieval with Compositional Cues}

\author{%
  Tingyu ~Lin, Robert ~Sablatnig\\
  Computer Vision Lab, TU Wien\\
  1040 Vienna, Austria \\
  \texttt{\{tylin, sab\}@cvl.tuwien.ac.at} \\
}

\begin{document}

\maketitle

\begin{abstract}

  In analyzing vast amounts of digitally stored historical image data, existing content-based retrieval methods often overlook significant non-semantic information, limiting their effectiveness for flexible exploration across varied themes. To broaden the applicability of image retrieval methods for diverse purposes and uncover more general patterns, we innovatively introduce a crucial factor from computational aesthetics, namely image composition, into this topic. By explicitly integrating composition-related information extracted by CNN into the designed retrieval model, our method considers both the image's composition rules and semantic information. Qualitative and quantitative experiments demonstrate that the image retrieval network guided by composition information outperforms those relying solely on content information, facilitating the identification of images in databases closer to the target image in human perception. Please visit https://github.com/linty5/CCBIR to try our codes.
\end{abstract}

\section{Introduction}

With the advancements in digital technology and the increasing emphasis on preserving historical records, a wealth of cultural heritage materials has undergone digitization. Paper-based repositories have undergone expansion and transformation into digital repositories, thus providing foundational resources for humanities scholars and fostering the development of related analytical techniques \cite{crane2003towards}, \cite{sinn2012impact}, \cite{kleber2013cvl}. Consequently, numerous historical image databases and datasets have come into existence. However, manually searching for similar images in extensive multimedia archives is impractical, making automated image and video retrieval systems crucial for accessing and analyzing historical image archives \cite{muhling2019content}. Existing popular image retrieval methods, primarily based on semantic or content-based matching, often utilize deep learning techniques to enhance feature extraction, focusing on multimodal retrieval and improvements in retrieval speed and security \cite{gabeur2020multi}, \cite{yan2020secure}. Meanwhile, image retrieval regarding photographic settings that determine the quality and style of photographs has received much less attention, resulting in the underutilization of information beyond semantics.

One frequently neglected yet vital aspect is compositional cues. Humans quickly recognize composition rules, which are extensively utilized by photographers \cite{liu2010optimizing}. Composition rules aid in conveying structural information, thus enhancing the effectiveness of image retrieval \cite{lee2018photographic}. Within photography, the quality assessment includes physical image parameters, such as size, aspect ratio, color depth, and higher-level perceptual aspects. This latter category, computational aesthetics, deals with the rules of composition, color, clarity, and semantics, offering a more nuanced understanding of image quality \cite{datta2008image}, \cite{debnath2020computational}. Studies on aesthetic features for similarity detection have focused on color composition \cite{ha2020color} and sketch-based composition cues \cite{sampaio2023scene}. The grayscale nature of most historical images, lacking color information, emphasizes the importance of compositional cues in historical image retrieval. More research is needed to directly extract composition features from images, particularly historical ones, for matching purposes. Combining semantic content and composition for retrieval would better assist historians and photography experts in analyzing historical materials, revealing the quality and potential intentions behind photos and films from various perspectives.

To tackle this challenge, we introduce a novel image retrieval approach that synergizes composition and content features, augmented by a specialized training and evaluation pipeline leveraging historical footage. Our method consists of two primary components: a composition feature extraction network and a content retrieval network. The composition network named Composition Clues Network (CCNet) is inspired by the composition branch from C. Hong et al. \cite{hong2021composing}'s Composition-Aware Cropping Network (CACNet), which assumes that composition rules can be learned and explicitly modeled within the network to guide image cropping effectively. We optimized this composition branch to serve as our image composition information extractor. Trained on the KU-PCP dataset \cite{lee2018photographic}, it extracts Class Activation Maps (CAMs) \cite{zhou2016learning} to encode the Key Composition Map (KCM), which is then passed to the image retrieval network to guide the training and retrieval of content features.

Our proposed Content-Based Image Retrieval Network (CBIRNet) merges composition information with content feature extraction. It was trained and tested using selected images and annotations from the publicly available HISTORIAN dataset \cite{helm2022historian}, a richly annotated historical video collection that offers annotations for over-scanned areas, start and end times of shots in videos, and shot types. Experimental results demonstrate that our CBIRNet, leveraging both composition and content information, can find images that are perceptually closer to the target image across various styles compared to networks relying solely on content-based retrieval.

\section{Related Work}

This section outlines the developmental trajectory and influential approaches in image composition analysis and content-based image retrieval.

\subsection{Image Composition Analysis}

In photography, image composition refers to the arrangement of elements within the frame, guiding the viewer's attention to the photographer's intended focus. Following artistic principles, this arrangement signifies the harmony of visual elements and is considered a critical factor in assessing aesthetic quality \cite{lee2018photographic}, \cite{liu2020composition}. Current methods for identifying composition rules involve computational feature design, such as calculating the distance between the subject's center and four centroids to detect the rule of thirds, measuring the dominance of diagonal lines, and assessing visual weight balance by the areas of two regions in the golden ratio, among other complex features \cite{bhattacharya2010framework}, \cite{tang2013content}.

While image composition analysis involves subjective elements, it can still be systematically categorized using established photographic principles. Considerable research, especially using CNNs, has been devoted to composition classification. For instance, T. Lee et al. \cite{lee2018photographic}'s work combines CNN-extracted features with composition rules and includes a sky detector for photographic composition classification and element detection. Similarly, C. Hong et al. \cite{hong2021composing} introduced a composition branch utilizing CNN features and CAMs to form KCM, providing composition information for downstream tasks. Previous efforts have explored conventional and CNN-based approaches within computational aesthetics, with our work further refining these methods and expanding their application domains.

\subsection{Content-Based Image Retrieval}

Content-Based Image Retrieval (CBIR) is generally categorized into two main tasks: Category-level Image Retrieval (CIR) and Instance-level Image Retrieval (IIR). CIR aims to retrieve images in the same category as the query image, while IIR targets images with the exact instance depicted in the query \cite{chen2022deep}. This paper concentrates on retrieving historical images that present objects of the same category with a similar layout. The typical CBIR system process is segmented into online and offline stages: "online" refers to operations on the query target image, while "offline" refers to processing the database for the query. Once a target image is an input, its feature vector is extracted and then searched, scored, and ranked against feature vectors from the database images. The results are finally returned and reordered according to similarity \cite{li2021recent}. We primarily focus on image representation within this workflow, namely extracting image features.

In CBIR systems, feature extraction methods mainly include conventional and CNN-based approaches. Conventional strategies are further classified as global and local feature extraction. Global attributes encompass color, shape, texture, and structure, allowing combinations across different features \cite{li2005generative}, \cite{wang2014content}, \cite{wang2011interactive}. Local features are identified using renowned algorithms such as SIFT and its variations \cite{zhou2015bsift}, \cite{zhang2013edge}, along with codeword-based methods designed for database search efficiency, including Bag of Words (BoW) and its enhancements \cite{perronnin2010large}, \cite{jegou2010aggregating}. Despite the success of these techniques, CNN-based strategies tend to surpass them in performance. Within this domain, there are two primary strategies: one leveraging classification-trained features for retrieval tasks \cite{babenko2014neural}, and another applying deep metric learning for end-to-end training of retrieval systems \cite{gordo2017end}, \cite{wang2019multi}. The former focuses on strategically using selective search and weighting to apply features optimally, whereas the latter concentrates on refining the metric used for measuring similarities.

\section{Methodology}
This paper proposes a dual-network approach for image retrieval tasks incorporating compositional information. Comprising two unique network structures, the compositional and retrieval branches are trained using distinct datasets. In this section, we elaborate on the datasets used, describe the network designs, and provide insights into the implementation process.

\subsection{Dataset}
For the compositional branch of our method, a compositional classification dataset suffices for training. However, the situation is more complex for the retrieval task dataset. Identifying datasets with similar and dissimilar content and composition content is essential to validate historical image retrieval techniques based on content and composition information.

\subsubsection{Composition Dataset}

The image composition dataset we use is the KU-PCP dataset introduced by J.-T. Lee et al. \cite{lee2018photographic}. It is a photography composition dataset categorized by human annotation into nine labels: rule-of-thirds, center, horizontal, symmetric, diagonal, curved, vertical, triangle, and repeated pattern. The dataset comprises 4,251 outdoor photos, with 3,169 for training and 1,082 for testing, including 20\% of the data having multiple labels. The distribution of images across different categories in the dataset is not entirely balanced. For instance, the rule-of-thirds accounts for 22.6\% in the training set and 9.4\% in the test set. At the same time, the curved category comprises 5.9\% and 6.9\% of the training and testing sets, respectively. This disparity stems from the dataset creators' methodology, which applies different techniques across categories. For categories with varied styles, such as the rule of thirds, the dataset creators employed CNN-based methods, necessitating a larger sample size and increasing the number of samples for these categories within the dataset. Conversely, handcrafted detectors were used for some categories, resulting in fewer samples. This approach may pose challenges for reproducing or matching their benchmarks with different models and code implementations. In our study, we utilize a unified model to extract composition information. However, the skewed distribution affects our model's performance, making it less effective in some categories than others.

\subsubsection{Retrieval Dataset}

\begin{figure}[h]
    \centering
    \includegraphics[width=0.6\linewidth]{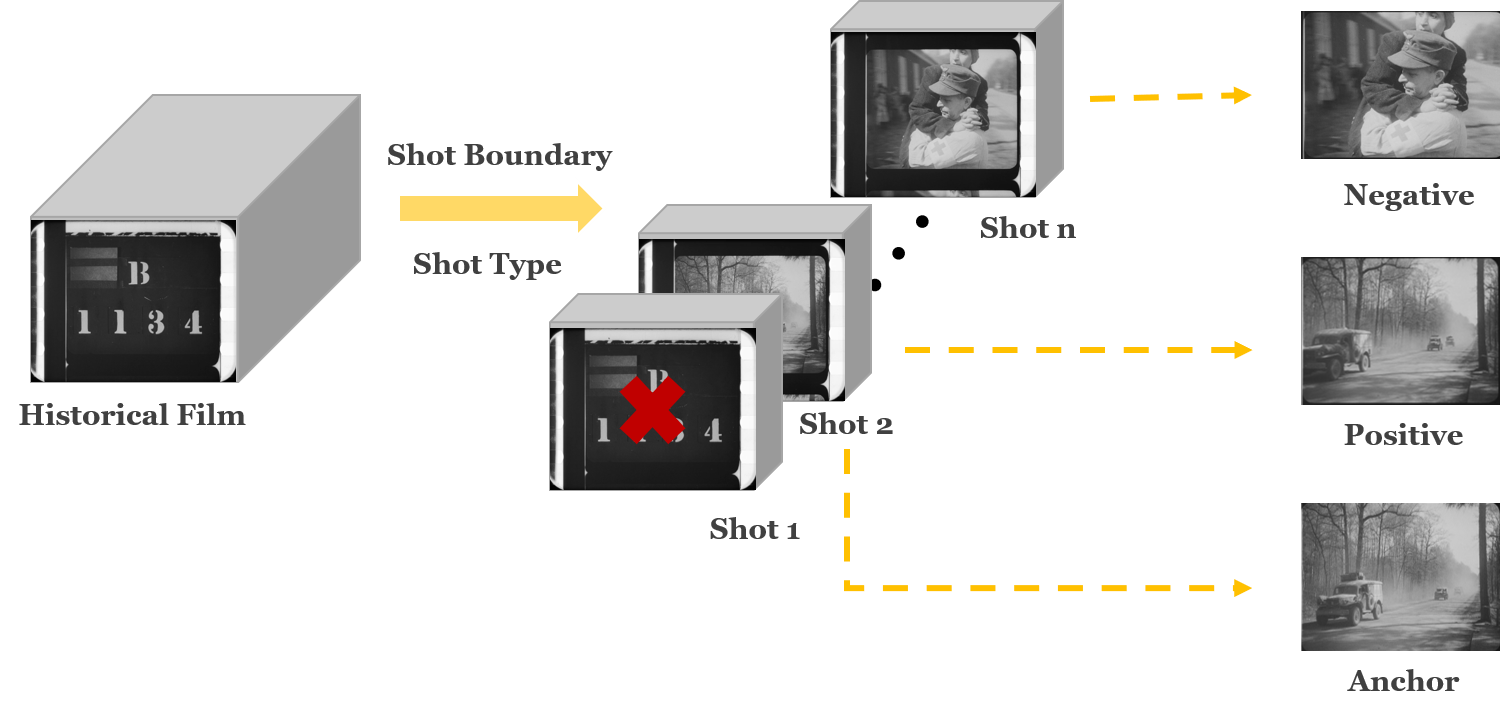}
    \caption{Illustration of converting historical films into pairs of images for retrieval experiments.}
    \label{Fig1}
\end{figure}

This paper presents an efficient approach for retrieval task data collection, assuming that images from adjacent frames within a shot share close content and compositional information, whereas images from different shots differ significantly. A "shot" is the primary film production unit in cinematography, encompassing a sequence of continuous frames. It is more extensive than a single frame but smaller than an entire scene. Shots within the same scene generally share the same theme, although camera settings may vary \cite{helm2022histshot}. For the process from historical video data to selecting anchor, positive, and negative samples, see Figure \ref{Fig1}. During the training and evaluating phase, an anchor image is selected from the dataset, and then the subsequent image within the same shot group is a positive sample. Conversely, a random frame from another shot is chosen as a negative sample. Despite negative samples and anchor images potentially sharing the same scene, their differing shot groups ensure notable variance in content and composition. Thus, employing shot boundary information allows for extracting image groups from films that are similar and dissimilar in content and composition. These groups can then be further refined by excluding specific shot types.

We utilized the HISTORIAN dataset proposed by D. Helm et al. \cite{helm2022historian}, encompassing 98 films annotated with 10,593 shots, complete with shot type and boundary details. From this comprehensive collection, images were strategically selected by capturing the first and last frames of every ten frames within each shot, with a maximum of seven images per shot. After excluding Intertitle (I) and Not Available/None (NA) shot types, our refined dataset consisted of 8,432 images from 33 films (6,778 for training and 1,654 for testing). These images span shot types such as Extreme Long Shot (ELS), Long Shot (LS), Medium Shot (MS), and Close Up (CU), reflecting the theme of liberating Nazi concentration camps during World War II. The dataset covers diverse categories, including portraits, architecture, vehicles, and natural landscapes, categorized by various compositional rules. Furthermore, images were cropped according to over-scanned area data to refine the focus on relevant visual information.

\subsection{Network Architecture}

\begin{figure}[h]
    \centering
    \includegraphics[width=0.8\linewidth]{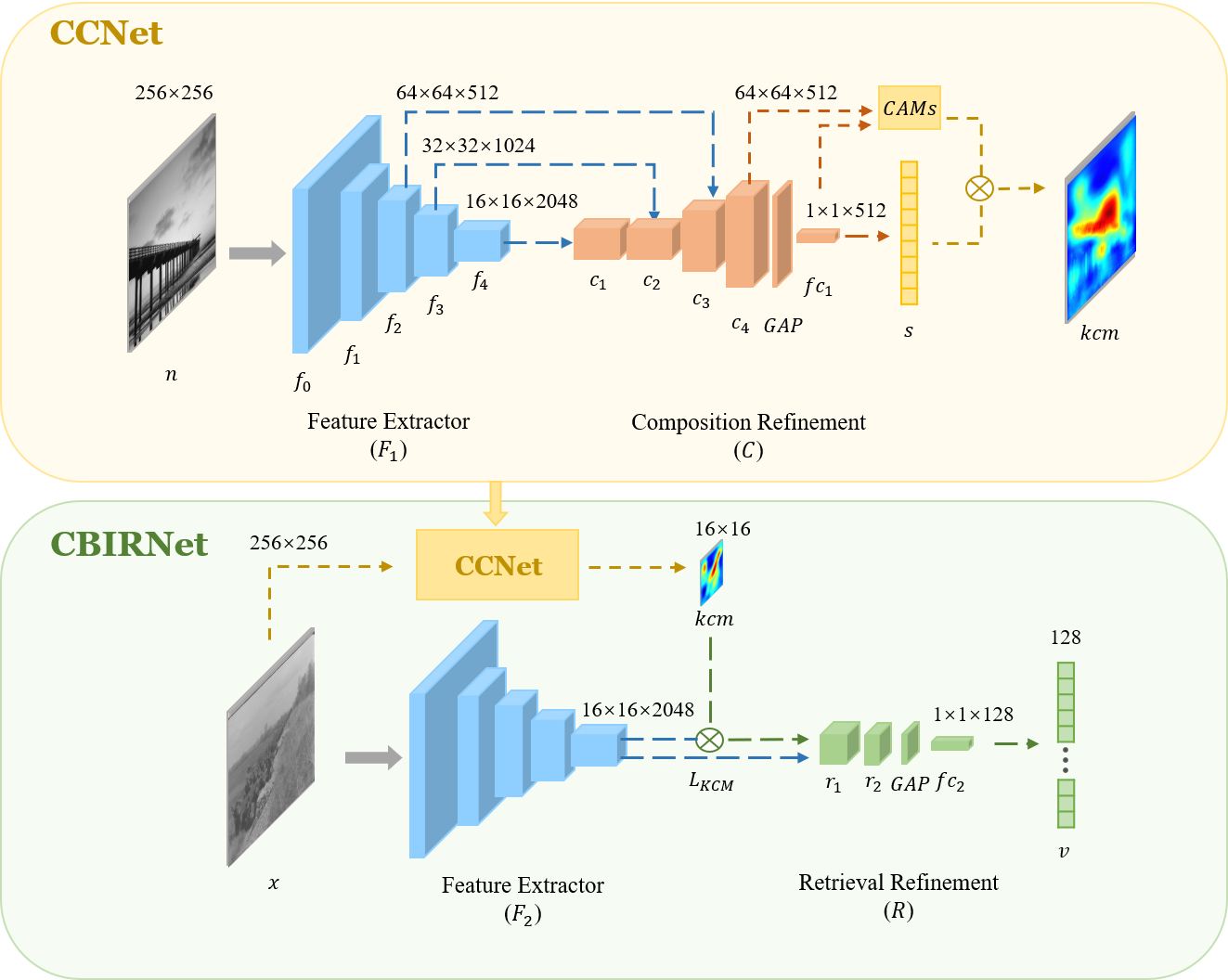}
    \caption{Illustration of our proposed CCNet and CBIRNet.}
    \label{Fig2}
\end{figure}

This section details the architecture of our proposed Composition Clues Network (CCNet) and Content-Based Image Retrieval Network (CBIRNet), explaining how they enhance historical image retrieval by considering both composition and content information. The overall architecture is illustrated in Figure \ref{Fig2}. 

CCNet is designed to extract composition-related cues from images, operating as a supervised classification model that categorizes composition types. It employs a pre-trained ResNet50 as its backbone architecture ($F_1$), enabling feature extraction from single-channel input images. $F_1$ consists of a preprocessing step ($f_0$) and four main components ($f_1$ to $f_4$), extracting progressively deeper features from grayscale images during training and testing stages. Features extracted by the $f_4$ component of $F_1$ are fed into a composition feature refining network ($C$), structured around four CNN layers ($c_1$ to $c_4$). Notably, features from $F_1$'s $f_2$ and $f_3$ layers are integrated into the $c_3$ and $c_2$ layers of $C$, respectively. This integration strategy enables $C$ to capture composition features across multiple scales effectively.

The features extracted by $C$ are aggregated and then classified through Global Average Pooling ($GAP$) and a fully connected layer ($fc_1$) to predict the composition type of the image. Additionally, CCNet employs the KCM mechanism from CACNet \cite{hong2021composing}, utilizing weighted aggregation of $CAMs$ from different composition types to provide an intuitive interpretation of image composition. Each $CAM$ specifically targets an activation map of a given category, with the category's predicted score mapped back to a preceding convolutional layer to generate the $CAM$. This process effectively highlights the decision-making regions associated with a specific category. By integrating the predicted scores $s$ for all categories with the $CAMs$ through a weighted fusion, we can pinpoint the regions of interest that the model relies on to make decisions in the task of image composition classification. Such integration highlights the areas within the image containing compositional information, resulting in $kcm$, representing a weighted map of composition information. Using this regional information to guide retrieval allows the retrieval model to focus on locations strongly related to compositional information. Consequently, by assessing whether the compositional areas of two images are consistent, composition is incorporated into the criteria for determining image similarity.

The other component, CBIRNet, consists mainly of a feature extractor $F_2$ and a retrieval feature refining network $R$, using the widely adopted ResNet50 as its backbone. Historical grayscale image $x$ is inputted into the network during training and testing. The features $F_2(x)$ extracted from $x$ by $F_2$ are fused with the $kcm$ obtained from inputting $x$ into CCNet, following a predefined scale $L_{KCM}$. $L_{KCM}$ is a ratio value ranging from 0 to 1, representing the proportion of $kcm$'s influence in the final fused feature map. For example, when $L_{KCM}$ is set to 0.8, the final feature map is obtained by taking 80\% of the input feature map multiplied by $L_{KCM}$ and adding it to 20\% of the input feature map. This process yields a feature map that emphasizes areas relevant to compositional cues, which is then dimensionally reduced through subsequent layers $r_1$ and $r_2$. Through a series of convolutions, global pooling ($GAP$), and a final fully connected layer ($fc_2$), the network transforms image features into a relatively compact feature vector $v$ that retains both content and composition information for subsequent image retrieval tasks.

\subsection{Implementation Details}

Our model is implemented within the PyTorch framework. Throughout the training process for both tasks, images are resized to 256 × 256 and normalized within a range from 0 to 1. For optimization, we employ the Adam optimizer, initializing the parameters of all networks using the Xavier method \cite{glorot2010understanding} and loading pre-trained model parameters for the backbone. CCNet and CBIRNet utilize distinct loss functions to optimize performance. CCNet employs a cross-entropy function to compute losses across various categories. Conversely, the training of CBIRNet employs a cosine embedding loss strategy, which calculates the losses between the anchor image and positive samples, as well as between the anchor image and negative samples, to balance the similarities and differences among images.

\section{Experiments}

Our experiment investigated whether the compositional information from the CCNet composition model ultimately aids the CBIRNet retrieval model in achieving better image retrieval results. The specific approach involved adjusting the influence parameter $L_{KCM}$ within CBIRNet to examine its impact. We evaluated the model's performance using multiple metrics, including custom-designed ones, to conduct a comparative assessment.

\subsection{Image Composition}

\begin{figure}[h]
    \centering
    \includegraphics[width=0.75\linewidth]{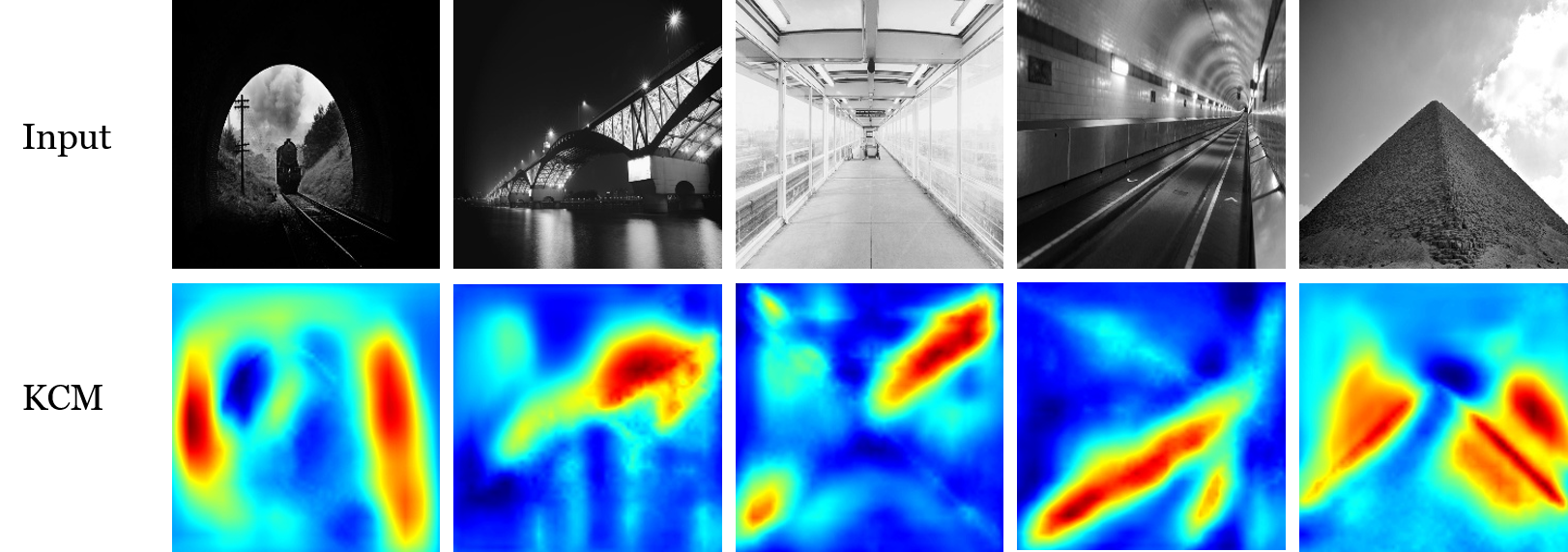}
    \caption{Visualisation of KCM effect. The top row features the original grayscale images, and the bottom row highlights the KCM, pinpointing key compositional areas as detected by our model.}
    \label{Fig3}
\end{figure}

For the image composition network, we evaluated our model on the grayscale KU-PCP dataset. Our model achieved an accuracy of 0.73, precision of 0.71, recall of 0.70, and an $F_1$ score of 0.70. Considering our task involves removing crucial color information, the significant reduction in information makes composition classification more challenging. We especially, given the dataset's unbalanced distribution and the high variability of samples, make it difficult for a relatively simple model to focus on compositional information. Figure \ref{Fig3} showcases the impact of our CCNet-trained model through visualizations of KCM. The model's ability to identify critical compositional areas across various compositional rules underscores its effectiveness. Given the accuracy and other metrics of CCNet, we are confident that it has significantly learned to capture compositional rules, making it a valuable tool for training subsequent CBIRNet models.

\subsection{Image Retrieval}

For CBIRNet, we fix the parameters of the CCNet obtained from previous training. By computing the cosine embedding loss between positive samples and the anchor and between negative samples and the anchor, we aim to reduce the distance between positive samples and the anchor while maximizing the distance between negative samples and the anchor.

\subsubsection{Quantitative Analysis}

\begin{figure}[h]
    \centering
    \includegraphics[width=0.9\linewidth]{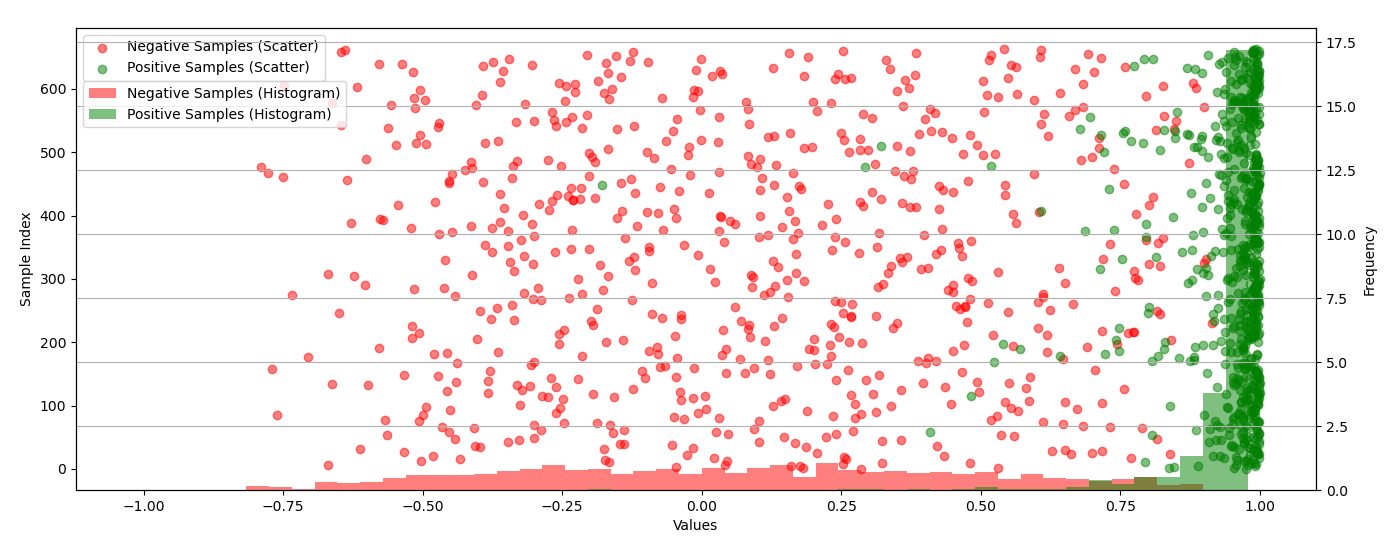}
    \caption{Scatter plot and histogram of positive and negative samples when $L_{KCM}$ is 0.5.}
    \label{Fig4}
\end{figure}

Two metrics are employed to assess model performance. The first metric is similar to the one used during training, where we calculate the cosine embedding loss between an anchor image with a positive sample and a negative sample to compare which model has the most negligible loss. The second metric calculates the cosine similarity between the anchor and two positive samples and between the anchor and two randomly chosen negative samples. Ideally, the cosine similarity between the anchor and positive samples should be higher than between the anchor and negative samples. When $L_{KCM}$ equals 0.5, the similarity distribution of all validation set samples with their corresponding four samples is shown in Figure \ref{Fig4}. This distribution suggests our model possesses a certain degree of discriminative ability, capable of generating varied scores based on the distinct characteristics of the samples.

A comparison of model metrics trained with varying $L_{KCM}$ values is shown in Table \ref{Tab1}. Besides calculating the average positive sample-to-anchor similarity and the average negative sample-to-anchor similarity, we propose a straightforward method of comparing these positive similarities against the negative ones. The model earns a point for each case where the similarity of a positive sample to the anchor surpasses that of a negative sample to the same anchor. The model's average similarity score is then determined by averaging these points across all anchor samples. In an ideal scenario, the optimal model would attain a score of 4, where scores nearing 4 indicate exceptional performance. This method provides a nuanced and equitable assessment of model efficacy. Models without added compositional information, except for the average similarity of negative samples, performed worse on other metrics than those augmented with KCM. 

\begin{table}[h]
  \caption{Effect of $L_{KCM}$ Values on Model Metrics}
  \label{Tab1}
  \centering
  \begin{tabular}{ccccc}
    \toprule
    $L_{KCM}$     & Cosine Embedding Loss      & Avg Pos Similarity       & Avg Neg Similarity     & Score \\
    \midrule
    0   & 0.2438 & 0.9423 & 0.0277  & 3.9759  \\
    0.5 & 0.2518 & 0.9561 & 0.0864  & 3.9819  \\
    0.8 & 0.2426 & 0.9480 & 0.0378  & 3.9820  \\
    \bottomrule
  \end{tabular}
\end{table}

\subsubsection{Qualitative Analysis}

\begin{figure}[h]
    \centering
    \includegraphics[width=0.8\linewidth]{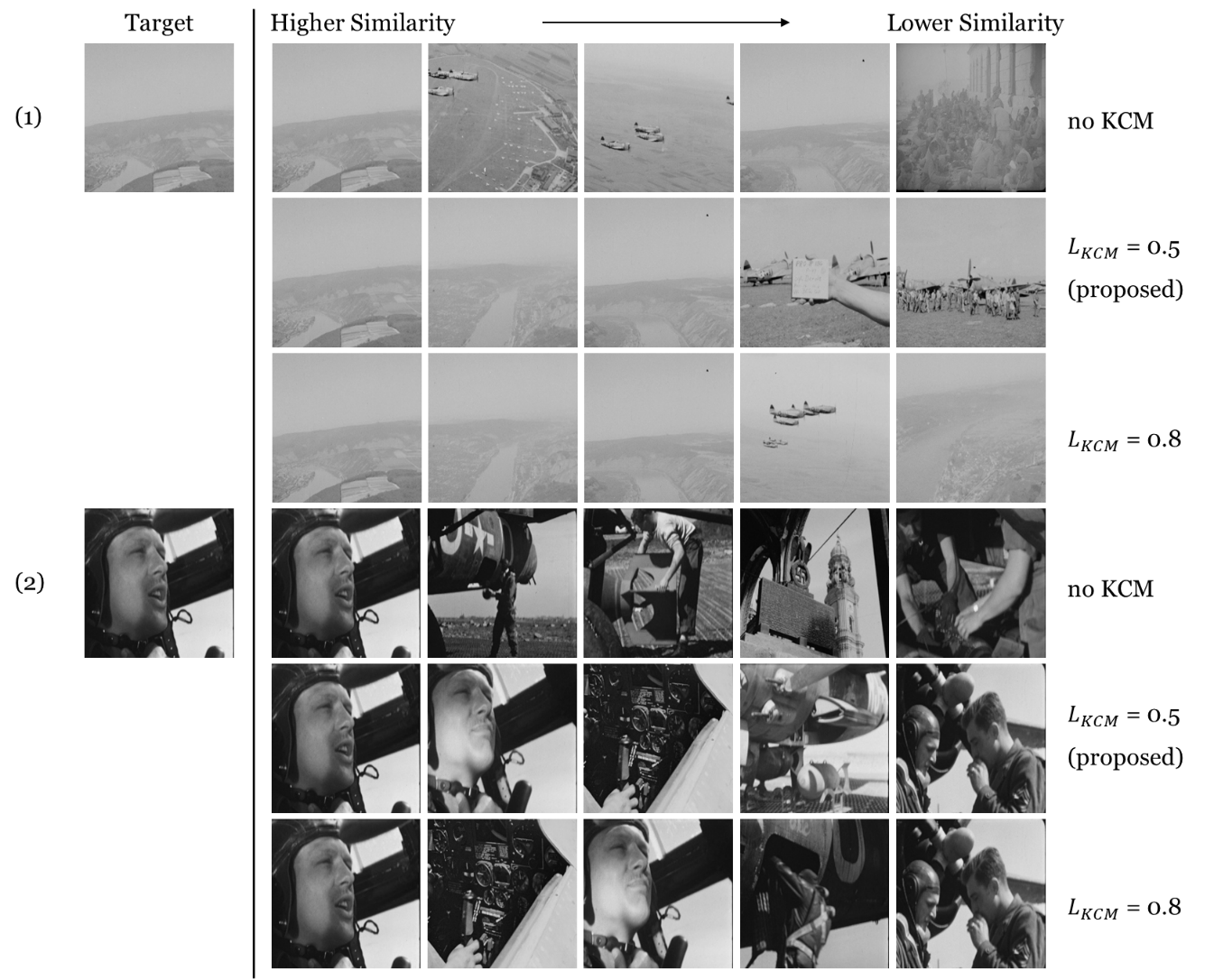}
    \caption{Comparison of retrieval results with different $L_{KCM}$. We selected only the central frame image from each shot in the test set as the target database for retrieval, returning the five highest similarity-scored images for a single image query.}
    \label{Fig5}
\end{figure}

As shown in Figure \ref{Fig5}, we present two sets of specific retrieval examples. Although all three models could identify the target image within the test sample set, indicating their ability to retrieve based on content, the model incorporating compositional information displayed a clear advantage when the alternative images came from different shots with slight variations.

In the first set of tests, models without compositional information, as shown in the first row, could identify some content-similar images, but it does not keep the composition consistent. The second and third rows integrated compositional information and consistently maintained the composition style vertically. In the second set of test samples, with $L_{KCM}$ set to 0.5, the model considered the compositional information's suggestion of areas of interest, focusing near the left-side facial area. This scale allowed the model to identify the most similar images that models without compositional information could not, placing them at the highest position after the original image. However, the third row, with a higher proportion of compositional information, resulted in insufficient content information, leading to worse outcomes. 

Integrating qualitative case studies with quantitative metric analyses demonstrates the superior performance of our model when utilizing an $L_{KCM}$ value of 0.5. Our approach effectively harnesses compositional elements and semantic content, yielding more precise and contextually relevant search outcomes in historical image databases. These findings underscore compositional information's crucial role and promising potential in this domain.

\section{Conclusion and Future Work}

This paper introduces an image retrieval method that integrates compositional cues and semantic information, utilizing a training and testing pipeline designed around historical images. Our compositional cues network and retrieval network were trained on the KU-PCP composition dataset and the HISTORIAN historical video dataset, respectively, and jointly tested to evaluate the impact of compositional information on retrieval outcomes. We transformed the training of our composition feature extractor into an image composition classification task. Despite needing to convert the KU-PCP Composition dataset into grayscale images to fit our historical image task and the dataset's class imbalance, our image composition classification model still achieved an accuracy of 0.73. This accuracy confirms its effectiveness in extracting compositional information from grayscale images. In retrieval tasks, models incorporating compositional information outperformed those relying solely on content-based methods, both quantitatively and qualitatively. Based on the distances between the features of positive and negative samples, our evaluation metrics suggest that models integrating compositional information have significant potential. 

However, to further validate the effectiveness and rationale of our approach, we need to design more experiments and optimize our current model from multiple perspectives. Specifically, to prove that the model can indeed identify images that are closer in both composition and content, constructing a dataset with image pairs that are either similar or dissimilar in composition and content is essential. In this paper, we employed a convenient method of extracting frames from the same shot or different shots within the historical video dataset to construct image pairs. While this method efficiently generates a large amount of training data, it lacks specific important pairs, such as those with similar composition but dissimilar content, and vice versa, since the way we generate image pairs binds the similarity of composition and content. Similarly, since we are working on content-based image retrieval, our dataset also needs pairs where the composition and content category are similar, but the content entities are not. Therefore, although our data processing method is effective, the contribution of building a dedicated dataset is irreplaceable and can better evaluate our model from all aspects.

Regarding the model itself, we proposed a dual-network approach to extract both compositional and content information. Merging these pieces of information has become one of the critical challenges. In our work, direct fusion has proven beneficial for the outcome. However, a more sophisticated fusion method could convincingly allow compositional information to play as significant a role as possible. Thus, our future work on this topic will explore these directions, seeking better ways to extract different types of information from historical images.

\begin{ack}
This work was supported by the Austrian Science Fund (FWF) -- doc.funds.connect, under project grant no. DFH 37-N: "Visual Heritage: Visual Analytics and Computer Vision Meet Cultural Heritage.". 
\end{ack}



\small \printbibliography 


\end{document}